\documentclass[10pt,twocolumn,letterpaper]{article}

\usepackage{iccv}
\usepackage{times}
\usepackage{epsfig}
\usepackage{graphicx}
\usepackage{amsmath}
\usepackage{amssymb}
\usepackage[T1]{fontenc}

\usepackage[breaklinks=true,bookmarks=false]{hyperref}

\iccvfinalcopy 

\def\iccvPaperID{****} 

\ificcvfinal\fi

\begin{document}
\title{Looking Beyond What You See: An Empirical Analysis on Subgroup Intersectional Fairness for Multi-label Chest X-ray Classification Using Social Determinants of Racial Health Inequities}
\newcommand{\repeatthanks}{\textsuperscript{\thefootnote}}

\author{Dana Moukheiber$^{1}$\thanks{Equal contribution. } \,\, Saurabh Mahindre$^{2*}$ \,\, Lama Moukheiber$^{1}$ \,\, Mira Moukheiber$^{1}$ \,\, Mingchen Gao$^{2}$\\
\vspace{-3mm}
\\  
$^1$Massachusetts Institute of Technology  \,\, $^2$ University at Buffalo}


\maketitle
\ificcvfinal\thispagestyle{empty}\fi


\begin{abstract}
There has been significant progress in implementing deep learning models in disease diagnosis using chest X-rays. Despite these advancements, inherent biases in these models can lead to disparities in prediction accuracy across protected groups. In this study, we propose a framework to achieve accurate diagnostic outcomes and ensure fairness across intersectional groups in high-dimensional chest X-ray multi-label classification. Transcending traditional protected attributes, we consider complex interactions within social determinants, enabling a more granular benchmark and evaluation of fairness. We present a simple and robust method that involves retraining the last classification layer of pre-trained models using a balanced dataset across groups. Additionally, we account for fairness constraints and integrate class-balanced fine-tuning for multi-label settings. The evaluation of our method on the MIMIC-CXR dataset demonstrates that our framework achieves an optimal tradeoff between accuracy and fairness compared to baseline methods.

\let\thefootnote\relax
\footnotetext{\hspace{-1.6em}\textit{
\newline
\fontsize{9.5}{9.5}\selectfont Published at the ICCV 2023 Workshop on Computer Vision for Automated Medical Diagnosis. France, Paris. Copyright 2023 by the author(s).}}

\end{abstract}
\vspace{-2.5mm}
\section{Introduction}
Deep learning algorithms trained on large-scale medical imaging data have been increasingly employed in real-world health applications in the past few years due to their potential to improve diagnostic accuracy and patient outcomes \cite{moukheiber2022few,hayat2021multi,agu2021anaxnet,wang2023enabling,ji2021improving}. However, these models also have the potential to learn and exacerbate pre-existing biases in data, perpetuating stereotypical connections and leading to negative consequences for marginalized communities \cite{10.1093/ejcts/ezab422}. The advancements and rapid deployment of computer vision models in healthcare settings highlight the necessity of acknowledging the possibility of inherent biases and risks across protected patient attributes related to disease, including race and gender.

 The misclassification of race as a biological aspect, rather than a social construct, further exacerbates health disparities. 
 To gain a more comprehensive understanding and enhance patient care, it is important to explore the patient's contextual environment, as a patient's clinical profile offers only a limited glimpse of the factors impacting their health \cite{nazer2023bias, holmes2023strategies, waite2021narrowing}. By examining the broader social, economic, and environmental factors through the lens of social determinants of health (SDOH), we can better comprehend the intricate interplay between racial biases and other determinants of health, shedding light on the mechanisms that perpetuate inequities in healthcare \cite{celi2022sources}. 
 
Several studies have been conducted to evaluate subgroup robust methods for addressing fairness bias across various domains of images with real-world applications \cite{zong2022medfair, gulrajani2020search, koh2021wilds, wiles2021finegrained}.
These studies mostly benchmark bias and fairness algorithms for binary classification \cite{izmailov2022feature,yang2023change,jabbour2020deep,zhang2021empirical} focusing on basic protective attribute categories along single dimensions. However, there has been limited research investigating intersectional bias with multiple demographic dimensions \cite{zhang2022improving,seyyed2021underdiagnosis,zhang2021empirical,kokhlikyan2022bias}, and the integration of SDOH into these studies remains unexplored, primarily due to privacy concerns\cite{driessen2023sociodemographic,10.1145/3351095.3373154}.

Herein, we adopt a simple and cost-effective method to fine-tune multi-label classification models for intersectional fairness across multiple attributes. Specifically, we measure intersectional group fairness in chest X-rays by considering race and two (SDOHs) — health insurance and income, forming eight distinct intersectional groups. Our approach adapts current subgroup robust methods to a balanced sampled multi-attribute dataset and addresses fairness constraints and class imbalance for multi-label settings. This adaptation extends beyond previous fairness intersectionality studies, which primarily focus on binary classification and intersections of two attributes. When applied to the MIMIC-CXR multi-label dataset, we show that our method performs well not only in classification performance but also in fairness metrics, outperforming established baselines.

\section{Related Works}
There have been several works proposed to combat fairness bias across benchmarks and real-world datasets:

\textbf{Debiasing Algorithms}: 
Debiasing algorithms play a crucial role in enhancing performance across minority subpopulations, thereby improving accuracy uniformly across different groups. These strategies can be broadly categorized into two types: those that rely on attributes being explicitly labeled and available during training, and attribute-agnostic methods, which do not necessitate direct access to such attributes during the training process. Empirical Risk Minimization (ERM)\cite{vapnik1999overview} , a foundational concept in machine learning, aims to minimize the average error across all samples by offering a generalized framework that can be adapted to include more sophisticated debiasing techniques. Some of these advanced methods, such as upweighting poorly performing samples\cite{liu2021just, nam2020learning, lahoti2020fairness}, directly build upon the ERM framework to ensure equitable performance across diverse groups. For instance, the Deep Feature Re-weighting (DFR) technique\cite{izmailov2022feature} begins with training a model using ERM and subsequently refines it by retraining with a balanced sample set across groups, illustrating a practical application of debiasing algorithms. These methodologies are critically evaluated using a variety of fairness metrics, ensuring measurable improvements in algorithmic fairness.


\textbf{Fairness Metrics}: Bias mitigation algorithms are evaluated using various fairness metrics. Some well-known metrics in this context are demographic parity \cite{10.1145/2783258.2783311}, equalized odds, and equality of opportunity. Demographic parity involves comparing the average prediction score across different subgroups. Equality of Opportunity takes the label distribution into account and assesses the True Positive Rate (TPR) gap among different groups. Equalized Odds \cite{hardt2016equality} measures both the TPR and False Positive Rate (FPR) gaps across various groups.





\textbf{Pre-processing, In-processing, and Post-processing Fairness Techniques}: Several prior works on mitigating fairness bias \cite{du2020fairness, pmlr-v182-marcinkevics22a} have been proposed, including pre-processing techniques \cite{wang2019balanced, kamiran2012data, celis2020data, alabdulmohsin2022reduction}, train-time techniques \cite{kim2019multiaccuracy, kamishima2012fairness}, and post-processing techniques \cite{hardt2016equality, kamiran2012data, pleiss2017fairness, yang2020fairness,alabdulmohsin2021near,schrouff2022diagnosing}, which aim to mitigate fairness bias before, during, or after model training, respectively. However, as presented by \cite{cherepanova2021technical, sagawa2020investigation}, over-parameterized models overfit to fairness objectives, which is an acute problem, especially in high-stake clinical settings. In addition, as outlined by \cite{deng2022fifa}, there are inherent imbalances that extend beyond class labels to include sensitive attributes, posing a challenge to the generalizability of fairness properties in over-parameterized models like neural networks. This issue becomes particularly pronounced in large and challenging datasets, especially when the model tends to favor minority attributes. 

\textbf{Intersectional Fairness}: Intersectional biases occur when protected attributes interact with each other. Previous research on bias evaluation in chest X-ray imaging has primarily focused on examining protected attributes such as race, age, and gender as mutually exclusive categories with single dimensions \cite{zong2022medfair, larrazabal2020gender}. Yet, only a few studies have explored protected attributes as non-mutually exclusive categories with multiple dimensions. Such an approach is critical, given that the disparities affecting intersectional subgroups can be profound, highlighting the need to address the compounded biases within marginalized communities. \cite{seyyed2021underdiagnosis} evaluated a DenseNet model on chest X-rays and revealed lower true positive rates (TPRs) across four categories of protected attributes, namely gender, age, race, and insurance type. However, their evaluation only focused on insurance as a proxy for social determinants of health (SDOH). Here, we focus on equalized odds as a metric to facilitate comparisons with previous work on studying fairness in medical imaging \cite{yang2023change, zhang2022improving, zong2022medfair}. Our study goes beyond race, gender, and age, avoiding pre-defined proxies of SDOH. Instead, we use and incorporate social determinant attributes sourced from country and tract-level data to better understand the complex interactions between various SDOH factors for more granular benchmarking and evaluation of fairness in clinical settings.

\section{Methods}
This section describes our methods and details the implementation process.
We assume we have a dataset of $n$ chest X-ray samples and $C$ target binary classes. Each sample can belong to one or more classes due to the multi-label nature of chest X-rays. The binary variable $y_{ik}$ denotes whether sample $i$ has class $k$ as positive, where $k \in C$. For intersectional groups, we define a set denoted as $G$. The binary variable $a_{ig}$ indicates whether a sample $i$ belongs to group $g$, where $g \in G$. We conduct experiments on MIMIC-CXR, which is divided into 190,000 training samples and 2,500 testing samples. We link MIMIC-CXR to MIMIC-IV and MIMIC-SDOH to create intersectional groups (see dataset details in Appendix \ref{appendix_dataset}). The number of samples used across eight intersectional groups is presented in Table \ref{tabel_samples}.

\begin{table}[!htp]\centering
\begin{center}
\caption
{Number of samples present across eight intersectional groups.}
\label{tabel_samples}
\hfill \break
\begin{tabular}{|l|l|l|c|}
\hline
Income & Insurance & Race & No. Samples \\
\hline\hline
Low & Low & White & 20,638 \\
Low & Low & Non-White &10,650\\
Low & High & White &20,308\\
Low & High & Non-White &26,261\\
High & Low & White &50,499\\
High & Low & Non-White &9,666\\
High & High & White &13,214 \\
High & High & Non-White &5,261 \\
\hline\hline
Total& & & 193,730 \\
\hline
\end{tabular}
\end{center}
\end{table}

\subsection{Pre-training a residual network for feature extraction}
We first train a neural network to extract features from chest X-ray images using the training data set. We adopt a residual network architecture as the feature extractor similar to prior studies on chest X-ray classification \cite{chauhan2020joint, ji2021improving,moukheiber2022few,10.1007/978-3-030-87589-3_12}. ERM is a standard training method that minimizes the average loss across samples. It has demonstrated its effectiveness in extracting meaningful features for both single and multi-label chest X-ray classification, and it has been incorporated into diverse fairness benchmarks \cite{liu2021just, nam2020learning, izmailov2022feature}. Consequently, we train the residual network via ERM. Since our training data set exhibits class imbalance, we use class weights, $p_k$, for each class for effective pre-training. The weighted version of binary cross-entropy loss used in pre-training is described in \ref{eq:bce_loss},
\vspace{1mm}
\begin{equation}\label{eq:bce_loss}
\begin{split}
L_{BCE}  = &- \sum\limits_{\substack{{i=1 ... n}}} \sum\limits_{\substack{{k=1 ... |C|}}} p_k y_{ik} \log{\hat{y_{ik}}} \\
 & + (1-y_{ik}) \log{(1-\hat{y_{ik}})} ,
\end{split}
\end{equation}
\vspace{1mm}

\noindent where $p_k = \frac{\text{\# of samples with no findings class}}{\text{\# of samples in class } k}$ is the positive weight for each class $k$ and \(\hat{y_{ik}}\) represents the predicted probability of class $k$ for the $i$-th sample, obtained after applying the sigmoid activation function to the logits.

\begin{table*}[!htp]\centering
\begin{center}
\caption{Model performance on MIMIC-CXR for multi-label classification incorporating the equalized odds difference fairness constraint. We use demographic attributes from MIMIC and social determinant attributes from MIMIC-SDOH dataset to form the intersectional groups. Averaged values are reported over 100 random trials.}
\label{table_results}
\hfill \break \break
\begin{tabular}{|l|c|c|c|c|}
\hline
Method & AUC$_{avg}$ ($\uparrow$) & EO\_Diff$_{avg}$ ($\downarrow$) & WACC$_{avg}$ ($\uparrow$) & AF$_{avg}$ ($\uparrow$)\\
\hline\hline
ERM &\textbf{0.7861} &0.4243 & 0.6438 &0.2195 \\
Fine-tuning  &0.7800 &0.3811 &0.6292 & 0.2481 \\
DFR &0.7806 &0.3677 &\textbf{0.6526} &0.2579\\
Fair Class-balanced Fine-tuning (Ours)  &0.7763 & \textbf{0.3224} &0.6045 &\textbf{0.2820}\\
\hline
\end{tabular}
\end{center}
\end{table*}

\subsection{Class balanced fine-tuning on a sampled dataset with fairness constraints}
For fine-tuning, we freeze the feature extractor and re-train a new final classification layer on a sampled data set with balanced group distribution \cite{kirichenko2022last} for robustness across intersectional groups\footnote{The pre-trained model weights and classifier weights will be made available on HuggingFace.}. Next, we add fairness constraints based on the false positive rate, $fpr$, (\ref{eq:fpr}) and false negative rate, $fnr$, (\ref{eq:fnr}) \cite{mao2023last}  \cite{10.5555/3491440.3491755} to our overall loss function. Intersectional groups and multi-label samples pose a challenge since a sample can belong to multiple classes and one of many intersectional groups.
Compared to prior work \cite{manisha2018fnnc} which only considers two values of a sensitive attribute in single label settings, we propose to calculate $fpr_{kg}$ (\ref{eq:fpr_kg}) and $fnr_{kg}$ (\ref{eq:fnr_kg}) for each pair of class $k$ and group $g$ separately to accommodate intersectional groups in multi-label settings.

\begin{equation}\label{eq:bce_fair}
    L_{fairness} = L_{BCE} + \alpha (fpr + fnr)
\end{equation}

\begin{equation}\label{eq:fpr_kg}
    fpr_{kg} = \left|
    \frac{\sum\limits_{\substack{i}} \hat{y_{ik}}(1-y_{ik}) a_{ig}}{\sum\limits_{\substack{i}}a_{ig}} - 
    \frac{\sum\limits_{\substack{i}} \hat{y_{ik}} (1-y_{ik}) (1-a_{ig})}{\sum\limits_{\substack{i}}(1-a_{ig})} \right|
\end{equation}
\vspace{1mm}
\begin{equation}\label{eq:fnr_kg}
    fnr_{kg} = \left|
    \frac{\sum\limits_{\substack{i}}  (1-\hat{y_{ik}})y_{ik}a_{ig}}{\sum\limits_{\substack{i}}a_{ig}} - 
    \frac{\sum\limits_{\substack{i}} (1-\hat{y_{ik}})y_{ik} (1-a_{ig})}{\sum\limits_{\substack{i}}(1-a_{ig})} \right|
\end{equation}
\vspace{1mm}

To account for class imbalance, we propose to use a weighted average of $fpr_{kg}$  and $fnr_{kg}$ for all $k \in C$ while aggregating for group $g$. The weights $w_k$ (\ref{eq:W}) are devised based on frequency $n_k$ of the positive label $y_k$ in the training set of $n$ samples. The final loss in the fine-tuning step is (\ref{eq:bce_fair}).

\begin{equation}\label{eq:W}
    w_k = \frac{(n - n_k)}{n}, 
    W = \sum\limits_{k} w_k
\end{equation}

\begin{equation}\label{eq:fpr} 
    fpr = \frac{1}{|G|}\sum\limits_{\substack{g}}\frac{1}{W}\sum\limits_{\substack{k}} fpr_{kg} w_k  
\end{equation}

\begin{equation}\label{eq:fnr} 
    fnr = \frac{1}{|G|}\sum\limits_{\substack{g}}\frac{1}{W}\sum\limits_{\substack{k}}fnr_{kg} w_k
\end{equation}

\newcommand{\probP}{\text{I\kern-0.15em P}}

\vspace{2mm}

\noindent \textbf{Compared Methods}. We consider three baseline methods, followed by our proposed method:  
\vspace{2mm}

\noindent \textbf{ERM}: Training a residual network where the final layer is a classification layer without considering intersectional groups.  \\
\textbf {Fine-tuning}: Re-training a new final classification layer of the residual network using an imbalanced sampled dataset \cite{kirichenko2022last}. \\
\textbf{Deep Feature Reweighting (DFR)}: Re-training a new final classification layer of the residual network using a balanced sampled dataset across intersectional groups.\\
\textbf{Fair Class-balanced Fine-tuning (Ours)}: Re-training a new final classification layer of the residual network using a sampled dataset balanced across inter-sectional groups, considering both fairness constraints and class imbalance. \\

\noindent \textbf{Evaluation Metrics}.
To evaluate the performance of our model on the unseen test set, we use weighted accuracy (WACC) and Area under the ROC Curve (AUC) \cite{mao2023last}. We use two metrics for classification: WACC, which accounts for the class imbalance and AUC. For multi-label settings, these metrics are averaged across all labels. For fairness evaluation, we adopt two metrics: equalized odds difference (EO\_Diff) \cite{hardt2016equality} and accuracy-fairness (AF) \cite{mao2023last}. EO\_Diff calculates the maximum between two differences: the difference between the minimum and maximum true positive rates and the difference between the minimum and maximum false positive rates across all intersectional groups \cite{hardt2016equality} \cite{verma2018fairness}. For multi-label settings, we calculate EO\_Diff for each label separately and then average it as shown in \ref{eq:eq_odd_diff}.  
\begin{equation}
\begin{split}
   EO\_Diff_k = \max\{ \Delta tpr_k , \Delta fpr_k\}
\end{split}
\end{equation}
\begin{equation}\label{eq:eq_odd_diff}   
   EO\_Diff_{avg} = \frac{\sum\limits_{\substack{k}} EO\_Diff_k}{|C|}
\end{equation}
\vspace{1mm}

\noindent AF is a metric derived from an equal-weight linear combination of weighted accuracy and fairness: 
AF = WACC - EO\_Diff \cite{mao2023last}. 


\subsection{Implementation Details} 

\textbf{Pre-training details}:
For pre-training, we use Adam optimizer with a learning rate of $10^{-4}$ and weight decay of $10^{-4}$. The mini-batch size is set to 32 due to hardware limitations and we perform training on the whole training set for three epochs.

\textbf{Training details}:
In order to create a balanced data set for fine-tuning, we sample 2500 study-ids per group at random from the training data set. For fine-tuning, we use Adam optimizer with a learning rate of $5 $x$10^{-5}$, weight decay of $10^{-3}$, and batch size of 32. We set the value of $\alpha$ to 1 for simplicity and perform fine-tuning for one full epoch. 

\textbf{Evaluation}:
While computing the weighted accuracy, we use a 0.5 threshold to convert the output probabilities to predicted labels.

\section{Results and Discussion}
We conduct multi-label classification on MIMIC-CXR to predict the 14 base classes, evaluating our proposed method compared to three baseline methods. The results of the model performance, along with fairness metrics, are presented in Table \ref{table_results}. We assess several metrics, including AUC$_{avg}$, WACC$_{avg}$, and AF$_{avg}$, where higher values indicate better performance. Conversely, for EO\_Diff$_{avg}$, a smaller value is considered more desirable. Our findings reveal that ERM achieves the highest AUC and second-highest WACC values but the lowest fairness metrics. Additionally, we observe that fair class balanced fine-tuning results in the lowest EO\_Diff$_{avg}$, suggesting reduced disparities and biases in model predictions among intersectional groups. Overall, fair class balanced fine-tuning exhibits the most favorable fairness metrics and overall performance. This is indicated by the highest AF value, although the WACC$_{avg}$ and AUC$_{avg}$ are slightly reduced.





\section{Conclusion}
In this study, we introduce a framework aimed at promoting equitable representation across diverse intersectional groups in high-dimensional, multi-label chest X-ray classification. We adopt an intersectional multi-attribute fairness perspective, we consider complex interactions within sensitive attributes, going beyond traditional protected attributes to include social determinants of health such as insurance and income. Our methods involves retraining the final classification layer of pre-trained models using a balanced sampled multi-attribute dataset. We also consider both fairness constraints in our overall loss and accommodate intersectional groups in multi-label settings, providing a simple and robust method for assessing fairness in real-world clinical applications. Our evaluation on the MIMIC-CXR dataset demonstrates that our method improves equalized odds difference and accuracy-fairness metrics marking a promising step forward in medical algorithms. 

\
\newline
\noindent\textbf{\large{Acknowledgements}\newline
}
\newline
D.M. and L.M. are supported by the National Institute of Biomedical Imaging and Bioengineering (NIBIB) under NIH  grant number NIH-R01-EB017205 and NIH-R01-EB030362. D.M. is supported by NIH National Library of Medicine (NLM) under contract number 75N97020C00013, and Massachusetts Life Sciences Center. M.G. is supported by NSF CAREER award IIS-2239537.

\clearpage

{\small
\bibliographystyle{ieee_fullname}
\bibliography{egbib}
}

\clearpage
\onecolumn
\section{Appendix}



\subsection{Dataset Description}
\label{appendix_dataset}
\noindent We utilize MIMIC-IV and MIMIC-SDOH to create attribute intersectional groups, which are then matched to the MIMIC-CXR images.

\begin{itemize}
    \item\textbf{MIMIC-CXR}: A comprehensive collection of de-identified Chest X-ray (CXR) data acquired from Beth Israel Deaconess Medical Center in Boston, United States. MIMIC-CXR includes 14 binary labels that indicate the presence or absence of pathology.\cite{johnson2019mimic}. Specifically, we follow \cite{moukheiber2022few} to resize the images into 2048 X 2048 and we only use anterior-posterior (AP) and posterior-anterior (PA) radiographs view positions. 

    \item \textbf{MIMIC-IV}: An electronic health records database that contains data on patients admitted to the intensive care unit at Beth Israel Deaconess Medical Center. This extensive database comprises comprehensive information about patients' clinical records, demographic details, laboratory findings, and other pertinent medical data \cite{johnson2023mimic}. In our study, we extract racial information from MIMIC-IV and categorize the race into two groups: white and non-white.

    \item \textbf{MIMIC-SDOH}: A database resulting from the integration of the MIMIC-IV clinical database with  Social Determinants of Health (SDOH) databases: County Health Rankings (CHR), Social Vulnerability Index (SVI), and Social Determinants of Health Database (SDOHD) \cite{yang2023evaluating}. Here, we focus on the variables available in the SDOHD \cite{ahrqwebsite}  which offers more detailed and granular SDOH data. 
    The SDOHD provides information at the county, census tract, and ZIP code levels, encompassing variables related to economic, healthcare, education, and social contexts as well as physical infrastructure. From this data, we extract health insurance information at the county level, focusing on the estimated percentage of the uninsured population for all income levels (under 65 years). Additionally, we gather income information at the tract level, specifically the median household income (dollars, inflation-adjusted to the data file year). Subsequently, we categorize both the health insurance and income data into two distinct groups each: high and low income, as well as high and low insurance coverage.
%

\end{itemize}





\end{document}